\pdfoutput=1

\documentclass[11pt]{article}

\PassOptionsToPackage{table}{xcolor}
\usepackage{array}
\usepackage{multirow} 

\usepackage[final]{acl}  

\usepackage[switch]{lineno} 

\usepackage{times}
\usepackage{latexsym}
\usepackage{amsmath}
\usepackage{subcaption}
\usepackage{makecell}
\usepackage{xcolor}
\usepackage{amsfonts}
\usepackage{alltt}
\usepackage[skins,breakable]{tcolorbox}
\usepackage{graphicx}  
\usepackage[LTH,T1]{fontenc}

\usepackage[utf8]{inputenc}
\usepackage{CJKutf8}
\DeclareUnicodeCharacter{0E01}{\thaiKoKai\hskip0pt\relax}
\DeclareUnicodeCharacter{0E02}{\thaiKhoKhai\hskip0pt\relax}
\DeclareUnicodeCharacter{0E03}{\thaiKhoKhuat\hskip0pt\relax}
\DeclareUnicodeCharacter{0E04}{\thaiKhoKhwai\hskip0pt\relax}
\DeclareUnicodeCharacter{0E05}{\thaiKhoKhon\hskip0pt\relax}
\DeclareUnicodeCharacter{0E06}{\thaiKhoRakhang\hskip0pt\relax}
\DeclareUnicodeCharacter{0E07}{\thaiNgoNgu\hskip0pt\relax}
\DeclareUnicodeCharacter{0E08}{\thaiChoChan\hskip0pt\relax}
\DeclareUnicodeCharacter{0E09}{\thaiChoChing\hskip0pt\relax}
\DeclareUnicodeCharacter{0E0A}{\thaiChoChang\hskip0pt\relax}
\DeclareUnicodeCharacter{0E0B}{\thaiSoSo\hskip0pt\relax}
\DeclareUnicodeCharacter{0E0C}{\thaiChoChoe\hskip0pt\relax}
\DeclareUnicodeCharacter{0E0D}{\thaiYoYing\hskip0pt\relax}
\DeclareUnicodeCharacter{0E0E}{\thaiDoChada\hskip0pt\relax}
\DeclareUnicodeCharacter{0E0F}{\thaiToPatak\hskip0pt\relax}
\DeclareUnicodeCharacter{0E10}{\thaiThoThan\hskip0pt\relax}
\DeclareUnicodeCharacter{0E11}{\thaiThoNangmontho\hskip0pt\relax}
\DeclareUnicodeCharacter{0E12}{\thaiThoPhuthao\hskip0pt\relax}
\DeclareUnicodeCharacter{0E13}{\thaiNoNen\hskip0pt\relax}
\DeclareUnicodeCharacter{0E14}{\thaiDoDek\hskip0pt\relax}
\DeclareUnicodeCharacter{0E15}{\thaiToTao\hskip0pt\relax}
\DeclareUnicodeCharacter{0E16}{\thaiThoThung\hskip0pt\relax}
\DeclareUnicodeCharacter{0E17}{\thaiThoThahan\hskip0pt\relax}
\DeclareUnicodeCharacter{0E18}{\thaiThoThong\hskip0pt\relax}
\DeclareUnicodeCharacter{0E19}{\thaiNoNu\hskip0pt\relax}
\DeclareUnicodeCharacter{0E1A}{\thaiBoBaimai\hskip0pt\relax}
\DeclareUnicodeCharacter{0E1B}{\thaiPoPla\hskip0pt\relax}
\DeclareUnicodeCharacter{0E1C}{\thaiPhoPhung\hskip0pt\relax}
\DeclareUnicodeCharacter{0E1D}{\thaiFoFa\hskip0pt\relax}
\DeclareUnicodeCharacter{0E1E}{\thaiPhoPhan\hskip0pt\relax}
\DeclareUnicodeCharacter{0E1F}{\thaiFoFan\hskip0pt\relax}
\DeclareUnicodeCharacter{0E20}{\thaiPhoSamphao\hskip0pt\relax}
\DeclareUnicodeCharacter{0E21}{\thaiMoMa\hskip0pt\relax}
\DeclareUnicodeCharacter{0E22}{\thaiYoYak\hskip0pt\relax}
\DeclareUnicodeCharacter{0E23}{\thaiRoRua\hskip0pt\relax}
\DeclareUnicodeCharacter{0E24}{\thaiRu\hskip0pt\relax}
\DeclareUnicodeCharacter{0E25}{\thaiLoLing\hskip0pt\relax}
\DeclareUnicodeCharacter{0E26}{\thaiLu\hskip0pt\relax}
\DeclareUnicodeCharacter{0E27}{\thaiWoWaen\hskip0pt\relax}
\DeclareUnicodeCharacter{0E28}{\thaiSoSala\hskip0pt\relax}
\DeclareUnicodeCharacter{0E29}{\thaiSoRusi\hskip0pt\relax}
\DeclareUnicodeCharacter{0E2A}{\thaiSoSua\hskip0pt\relax}
\DeclareUnicodeCharacter{0E2B}{\thaiHoHip\hskip0pt\relax}
\DeclareUnicodeCharacter{0E2C}{\thaiLoChula\hskip0pt\relax}
\DeclareUnicodeCharacter{0E2D}{\thaiOAng\hskip0pt\relax}
\DeclareUnicodeCharacter{0E2E}{\thaiHoNokhuk\hskip0pt\relax}
\DeclareUnicodeCharacter{0E2F}{\thaiPaiyannoi\hskip0pt\relax}
\DeclareUnicodeCharacter{0E30}{\thaiSaraA\hskip0pt\relax}
\DeclareUnicodeCharacter{0E31}{\thaiMaiHanakat\hskip0pt\relax}
\DeclareUnicodeCharacter{0E32}{\thaiSaraAa\hskip0pt\relax}
\DeclareUnicodeCharacter{0E33}{\thaiSaraAm\hskip0pt\relax}
\DeclareUnicodeCharacter{0E34}{\thaiSaraI\hskip0pt\relax}
\DeclareUnicodeCharacter{0E35}{\thaiSaraIi\hskip0pt\relax}
\DeclareUnicodeCharacter{0E36}{\thaiSaraUe\hskip0pt\relax}
\DeclareUnicodeCharacter{0E37}{\thaiSaraUee\hskip0pt\relax}
\DeclareUnicodeCharacter{0E38}{\thaiSaraU\hskip0pt\relax}
\DeclareUnicodeCharacter{0E39}{\thaiSaraUu\hskip0pt\relax}
\DeclareUnicodeCharacter{0E3A}{\thaiPhinthu\hskip0pt\relax}
\DeclareUnicodeCharacter{0E3F}{\textbaht\hskip0pt\relax}
\DeclareUnicodeCharacter{0E40}{\thaiSaraE\hskip0pt\relax}
\DeclareUnicodeCharacter{0E41}{\thaiSaraAe\hskip0pt\relax}
\DeclareUnicodeCharacter{0E42}{\thaiSaraO\hskip0pt\relax}
\DeclareUnicodeCharacter{0E43}{\thaiSaraAiMaimuan\hskip0pt\relax}
\DeclareUnicodeCharacter{0E44}{\thaiSaraAiMaimalai\hskip0pt\relax}
\DeclareUnicodeCharacter{0E45}{\thaiLakkhangyao\hskip0pt\relax}
\DeclareUnicodeCharacter{0E46}{\thaiMaiyamok\hskip0pt\relax}
\DeclareUnicodeCharacter{0E47}{\thaiMaitaikhu\hskip0pt\relax}
\DeclareUnicodeCharacter{0E48}{\thaiMaiEk\hskip0pt\relax}
\DeclareUnicodeCharacter{0E49}{\thaiMaiTho\hskip0pt\relax}
\DeclareUnicodeCharacter{0E4A}{\thaiMaiTri\hskip0pt\relax}
\DeclareUnicodeCharacter{0E4B}{\thaiMaiChattawa\hskip0pt\relax}
\DeclareUnicodeCharacter{0E4C}{\thaiThanthakhat\hskip0pt\relax}
\DeclareUnicodeCharacter{0E4D}{\thaiNikhahit\hskip0pt\relax}
\DeclareUnicodeCharacter{0E4E}{\thaiYamakkan\hskip0pt\relax}
\DeclareUnicodeCharacter{0E4F}{\thaiFongman\hskip0pt\relax}
\DeclareUnicodeCharacter{0E50}{\thaizero\hskip0pt\relax}
\DeclareUnicodeCharacter{0E51}{\thaione\hskip0pt\relax}
\DeclareUnicodeCharacter{0E52}{\thaitwo\hskip0pt\relax}
\DeclareUnicodeCharacter{0E53}{\thaithree\hskip0pt\relax}
\DeclareUnicodeCharacter{0E54}{\thaifour\hskip0pt\relax}
\DeclareUnicodeCharacter{0E55}{\thaifive\hskip0pt\relax}
\DeclareUnicodeCharacter{0E56}{\thaisix\hskip0pt\relax}
\DeclareUnicodeCharacter{0E57}{\thaiseven\hskip0pt\relax}
\DeclareUnicodeCharacter{0E58}{\thaieight\hskip0pt\relax}
\DeclareUnicodeCharacter{0E59}{\thainine\hskip0pt\relax}
\DeclareUnicodeCharacter{0E5A}{\thaiAngkhankhu\hskip0pt\relax}
\DeclareUnicodeCharacter{0E5B}{\thaiKhomut\hskip0pt\relax}

\usepackage{microtype}
\usepackage{arydshln} 
\usepackage{graphicx}
\usepackage{booktabs}     
\usepackage{multirow}     
\usepackage{makecell}     
\usepackage{array}        
\usepackage{placeins}     
\usepackage{arydshln}     

\definecolor{tred}{RGB}{251, 130, 132}
\definecolor{tcase}{RGB}{65, 151, 156}

\title{Source-Grounded Semantic Reinforcement Learning for Low-Resource Target-Language Generation}
\author{%
  \small
  \begin{tabular}[t]{@{}c@{}}
    Zeli Su\textsuperscript{1,2} \quad
    Ziyin Zhang\textsuperscript{3} \quad
    Zewei Pan\textsuperscript{3} \quad
    Zhou Liu\textsuperscript{4} \quad
    Dingcheng Huang\textsuperscript{5} \quad
    Dehan Li\textsuperscript{6} \\[3pt]
    Zhankai Xu\textsuperscript{2} \quad
    Longfei Zheng\textsuperscript{2} \quad
    Xiaolu Zhang\textsuperscript{2} \quad
    Jun Zhou\textsuperscript{2,\textdagger} \quad
    Wentao Zhang\textsuperscript{4,\textdagger}
  \end{tabular}
  \\[8pt]
  \footnotesize
  \begin{tabular}[t]{@{}c@{}}
    \textsuperscript{1} Minzu University of China \quad
    \textsuperscript{2} Ant Group \quad
    \textsuperscript{3} Shanghai Jiao Tong University \\[2pt]
    \textsuperscript{4} Peking University \quad
    \textsuperscript{5} Harbin Institute of Technology \quad
    \textsuperscript{6} South China University of Technology \\[3pt]
    \textsuperscript{\textdagger} Corresponding authors
  \end{tabular}
}

\begin{document}
\maketitle

\begin{abstract}
Low-resource target-language generation is often limited by scarce parallel data, while high-resource source-language monolingual data is abundant but difficult to use with standard supervised fine-tuning. We propose \textbf{Source-Grounded Semantic Reinforcement Learning (SG-SRL)}, a resource-utilization framework that converts source-language monolingual data into cross-lingual semantic supervision for target-language generation. SG-SRL performs reference-free reinforcement learning (RL) on source-language data using a cross-lingual semantic reward model, instantiated by a cross-lingual reranker that scores the semantic relevance between the source input and the target-language generation. While this induces severe verbosity-based reward hacking, a lightweight recovery stage using a small parallel corpus restores fluency, conciseness, and task format while preserving the semantic gains. Experiments on Chinese-to-Thai generation show that SG-SRL improves semantic grounding and factual coverage over cold-start SFT. Additional analyses on long-form transfer and Tibetan embedding-based rewards clarify the generalization behavior of SG-SRL and show that an encoder-based semantic reward can substitute for an LLM-based reranker in a realistic low-resource language setting.
\end{abstract}

\section{Introduction}

\begin{figure}[t]
    \centering
    \includegraphics[width=\columnwidth]{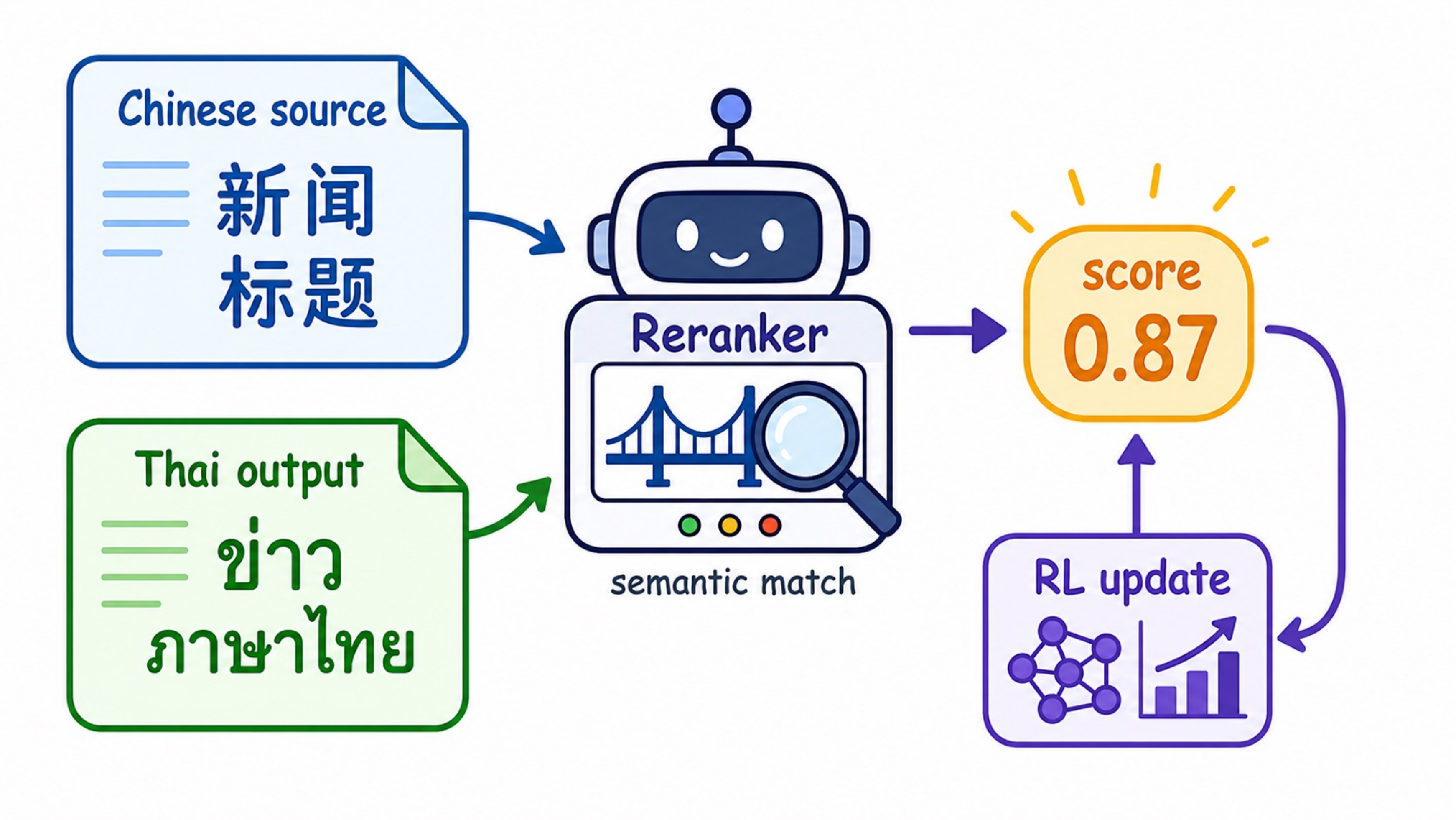}
    \caption{
    Reranker-based source-grounded semantic reward. A source-language input and a generated target-language output are treated as a cross-lingual query--candidate pair. The reranker estimates their semantic match and provides a scalar reward for RL without requiring a target-language reference.
    }
    \label{fig:reranker-reward}
    \vspace{-1em}
\end{figure}

Large language models (LLMs) have achieved strong general-purpose generation abilities through large-scale pretraining and post-training \citep{deepseekai2024deepseekv3, guo2025deepseekr1, yang2025qwen3}, but their performance remains highly uneven across languages. High-resource languages benefit from abundant pretraining text, instruction data, and task-specific supervision, whereas low-resource languages often suffer from unstable generation, hallucinated content, poor factual grounding, and weak task adaptation \citep{ustun2024aya, qin2025survey}. Continual pretraining on target-language corpora followed by supervised fine-tuning (SFT) on downstream tasks has been employed for extending LLMs to new languages \citep{wang-etal-2020-extending, joshi-etal-2025-adapting}, but it still relies on substantial target-language text or target-language task supervision, which is often unavailable for genuinely low-resource languages.

This data bottleneck becomes even more severe in cross-lingual generation. For many emerging or weakly supported target languages, high-quality parallel data is expensive to collect, and task-specific target-language annotations are even scarcer. In contrast, high-resource source-language data, such as Chinese or English news articles, is often abundant and reliable. Standard SFT cannot directly use such source-language monolingual data, because it requires a target-language reference for each training instance. This leads to the central question of this work: \emph{Can abundant source-language monolingual data be converted into useful supervision for low-resource target-language generation, without requiring target-language references during training?}

To this end, we need a supervision signal that can compare a source-language input with a target-language generation without a gold target-language reference. Multilingual rerankers provide a practical approximation: they score the relevance between a query and a candidate text \citep{nogueira-cho-2019-passage}, and recent LLM-based variants can distinguish semantically aligned cross-lingual pairs from unrelated ones \citep{sun-etal-2023-chatgpt, zhang-etal-2024-mgte, zhang-etal-2025-qwen3embedding}. As illustrated in Figure~\ref{fig:reranker-reward}, we treat the source input as the query and the generated target-language output as the candidate, and use a reranker as a source-grounded semantic reward. This converts source-language monolingual data into scalable RL supervision without target-language references.

Empirically, direct optimization of this reward exposes a semantic--form trade-off. The relevance-oriented reward encourages the policy to generate longer outputs that cover more source-side concepts, leading to \emph{verbosity-based reward hacking}: the intermediate RL checkpoint improves semantic coverage but becomes verbose, poorly formatted, and less fluent \citep{amodei2016concrete}. Stronger length or format constraints can reduce verbosity, but may also suppress useful source-to-target semantic learning. We therefore treat source-grounded semantic RL as \emph{semantic mid-training} rather than final optimization.

Based on this insight, we propose \textbf{Source-Grounded Semantic Reinforcement Learning (SG-SRL)}, a train--reinforce--recover framework for low-resource target-language generation. SG-SRL first learns target-language form from a small parallel corpus, then reinforces source-grounded semantics on source-language monolingual data, and finally reuses the parallel corpus to recover fluent and concise target-language generation. On Chinese-to-Thai generation, starting from SmolLM3-3B \cite{bakouch2025smollm3}, a model with weak Thai support, SG-SRL substantially improves over cold-start SFT, showing that semantic RL as mid-training can strengthen new-language semantic alignment. We further examine a Tibetan setting, where a strong LLM-based reranker is not available for the target language. By training an encoder-based Tibetan--Chinese embedding scorer, we demonstrate that SG-SRL can generalize to a realistic low-resource language scenario with a different form of relevance reward.

Our contributions are summarized as follows:
\vspace{-0.15cm}
\begin{itemize}
    \item We propose \textbf{SG-SRL}, a train--reinforce--recover framework that converts source-language monolingual data into semantic supervision to enable reference-free RL in the low-resource target-language generation setting.
    \vspace{-0.1cm}
    \item We instantiate SG-SRL with a multilingual reranker reward, and identify a semantic--form trade-off: direct optimization induces verbosity-based reward hacking, but the intermediate checkpoint can still learn useful cross-lingual semantic alignment.
    \vspace{-0.1cm}
    \item We show on Chinese-to-Thai generation with SmolLM3-3B that semantic RL as mid-training improves low-resource target-language generation over cold-start SFT, and we further verify language generalization in a Tibetan setting where an encoder-based embedding reward replaces the LLM-based reranker.
\end{itemize}

\section{Related Work}

\begin{figure*}[t]
    \centering
    \includegraphics[width=\textwidth]{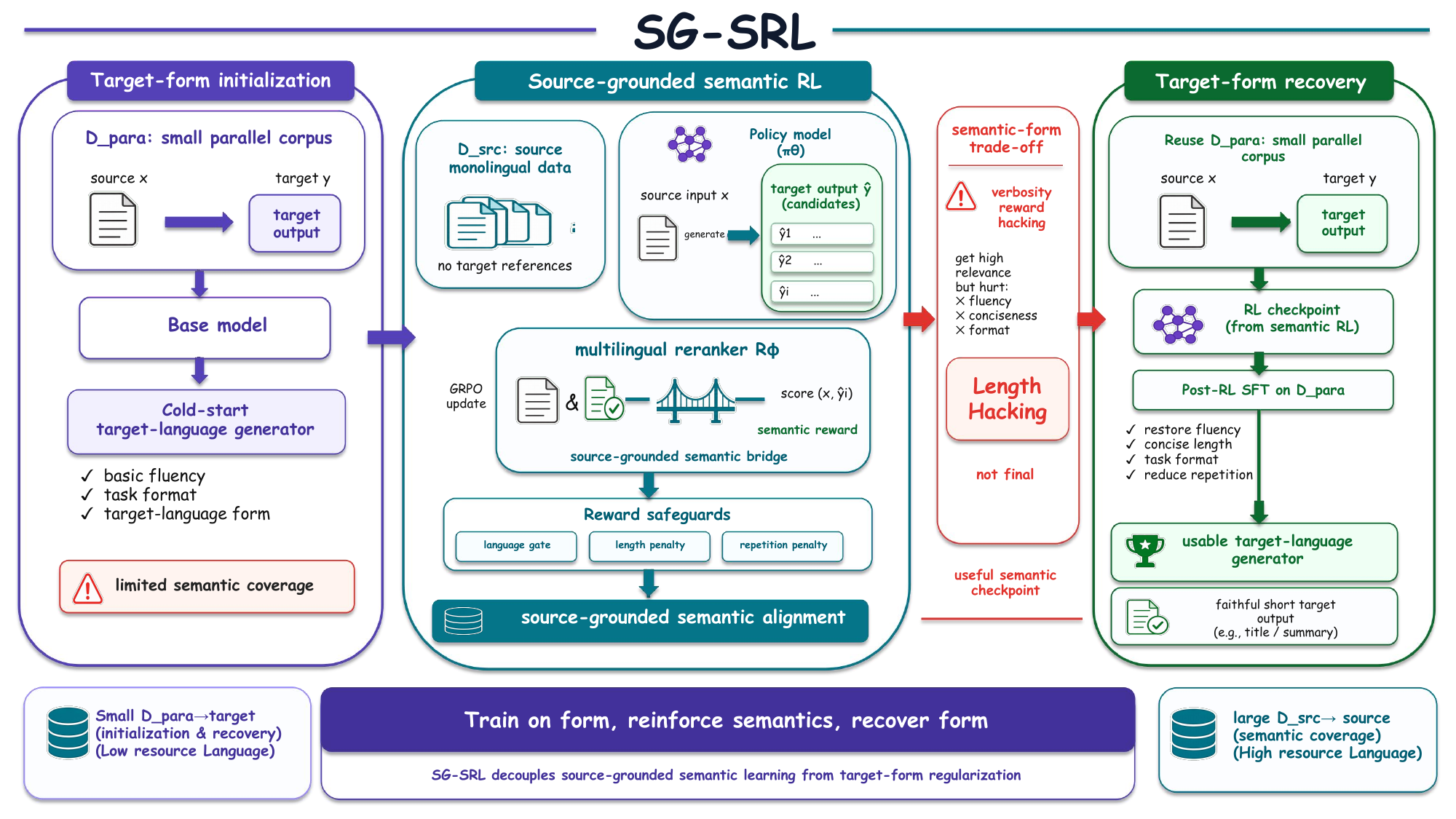}
    \caption{
    Overview of SG-SRL. The framework uses a small parallel corpus $\mathcal{D}_{\text{para}}$ for target-form initialization and target-form recovery, and a large source-language monolingual corpus $\mathcal{D}_{\text{src}}$ for source-grounded semantic RL. The RL stage uses a multilingual reranker as a weak semantic reward model, with language, length, and repetition safeguards. A final recovery stage reuses $\mathcal{D}_{\text{para}}$ to restore fluency, conciseness, and task format.
    }
    \label{fig:sgsrl-overview}
    \vspace{-1.4em}
\end{figure*}


\paragraph{Low-resource language expansion.}
Expanding language models to low-resource languages has commonly been approached through multilingual pretraining \citep{conneau-etal-2020-unsupervised,xue-etal-2021-mt5}, continued pretraining \citep{wang-etal-2020-extending,joshi-etal-2025-adapting}, or supervised fine-tuning on target-language data. However, these approaches still depend on the availability of target-language text or task-specific supervision. In genuinely low-resource settings, both pretraining corpora and high-quality downstream parallel data may be limited, motivating methods that can exploit abundant source-language data without requiring target-language references for every instance.

\paragraph{Semantic rewards for low-resource language learning.}
Recent work has explored reinforcement learning with semantic rewards as an alternative to token-level likelihood optimization for low-resource language expansion. \citet{su2026semanticrl} propose using embedding-level semantic rewards with GRPO to improve low-resource language capabilities while reducing alignment tax, showing that semantic-space optimization can preserve general capabilities better than conventional SFT.
Our work shares the same motivation, and further extends the horizon to a reference-free, cross-lingual generation setting.

\paragraph{Cross-lingual semantic matching and LLM-based reranking.} 
Dense multilingual representations and cross-lingual retrieval models provide a basis for measuring semantic correspondence across languages \citep{feng-etal-2022-language,bonifacio-etal-2021-mmarco}. Recent LLM-based rerankers further formulate relevance estimation as an instruction-following judgment over a query--candidate pair, often deriving a score from the model's preference for an affirmative answer \citep{sun-etal-2023-chatgpt,zhang-etal-2024-mgte,zhang-etal-2025-qwen3embedding}. We adopt this formulation by treating the source-language input as the query and the generated target-language output as the candidate. Unlike standard retrieval, however, we use the reranker's yes/no relevance probability as an online reward for policy optimization. This turns the reranker from an offline ranking module into an optimization target, which exposes relevance-oriented reward exploitation such as verbosity-based reward hacking.

\vspace{-0.05cm}
\paragraph{Reinforcement learning and reward hacking.}
Reinforcement learning has become a central tool for adapting language models to human preferences and task-specific objectives \citep{ouyang2022training,shao2024deepseekmath}. However, optimizing imperfect proxy rewards can lead to reward hacking, where models satisfy the literal reward function while violating the intended behavior \citep{amodei2016concrete}. In our setting, the reranker reward is semantically informative but relevance-oriented: longer generations can cover more source-side information and thus receive higher scores, even when they degrade fluency, conciseness, or task format.
Rather than treating this as a failure of RL alone, we show that the hacked intermediate policy can still acquire useful cross-lingual semantic alignment.

\section{Method}
\label{sec:method}

SG-SRL uses two data sources for different purposes: a small parallel corpus $\mathcal{D}_{\text{para}}$ for target-form initialization and recovery, and a large source-language monolingual corpus $\mathcal{D}_{\text{src}}$ for semantic mid-training. As shown in Figure~\ref{fig:sgsrl-overview}, SG-SRL follows a train--reinforce--recover procedure: train target-language form from $\mathcal{D}_{\text{para}}$, reinforce source-grounded semantics from $\mathcal{D}_{\text{src}}$, and recover target-language form again with $\mathcal{D}_{\text{para}}$.

\subsection{Problem Formulation}

We consider a high-resource source language $\mathcal{X}$ and a low-resource or weakly supported target language $\mathcal{Y}$. We are given a small parallel corpus
\begin{equation}
    \mathcal{D}_{\text{para}} = \{(x_i, y_i)\}_{i=1}^{N},
\end{equation}
where $x_i \in \mathcal{X}$ and $y_i \in \mathcal{Y}$, and a much larger source-language monolingual corpus
\begin{equation}
    \mathcal{D}_{\text{src}} = \{x_j\}_{j=1}^{M}, \quad M \gg N,
\end{equation}
which contains no target-language references.

The goal is to train a policy model $\pi_\theta$ that generates a target-language output $\hat{y} \in \mathcal{Y}$ conditioned on a source-language input $x \in \mathcal{X}$. Standard supervised fine-tuning only uses $\mathcal{D}_{\text{para}}$. SG-SRL converts $\mathcal{D}_{\text{src}}$ into RL training data by using a cross-lingual semantic scorer as a weak reward model.

\subsection{SG-SRL Overview}

SG-SRL has three stages:
\begin{enumerate}
    \item \textbf{Target-form initialization}: fine-tune the base model on $\mathcal{D}_{\text{para}}$ for 3 epochs to obtain a cold-start target-language generator.
    \item \textbf{Source-grounded semantic RL}: optimize the initialized model on $\mathcal{D}_{\text{src}}$ for 2 epochs using a reranker-based semantic reward.
    \item \textbf{Target-form recovery}: fine-tune the RL checkpoint again on $\mathcal{D}_{\text{para}}$ for 1 epoch to restore fluency, length control, and task format.
\end{enumerate}

The design separates target-language form learning from source-grounded semantic learning. The small parallel corpus gives reliable but limited form supervision, while the source-language monolingual corpus provides broader semantic coverage without target-language references.

\subsection{Target-form Initialization}

We first perform supervised fine-tuning on $\mathcal{D}_{\text{para}}$:
\begin{equation}
    \theta_{\text{sft}} =
    \arg\min_{\theta}
    - \mathbb{E}_{(x,y)\sim \mathcal{D}_{\text{para}}}
    \log \pi_{\theta}(y \mid x).
\end{equation}
This stage teaches the model basic target-language form, task format, and local generation style. However, because $\mathcal{D}_{\text{para}}$ is small, the resulting cold-start model may still miss source-side entities, events, or factual relations. The next stage therefore uses $\mathcal{D}_{\text{src}}$ to inject additional semantic grounding.

\subsection{Source-grounded Semantic RL}
\label{sec:source-grounded-semantic-rl}

Starting from $\pi_{\theta_{\text{sft}}}$, we perform RL on $\mathcal{D}_{\text{src}}$. For each source input $x \sim \mathcal{D}_{\text{src}}$, the policy samples
\begin{equation}
    \hat{y} \sim \pi_\theta(\cdot \mid x).
\end{equation}
Since no target-language reference is available, we use the reranker format in Figure~\ref{fig:reranker-reward}: the source input is treated as the query, and the generated target-language output is treated as the candidate document.

\paragraph{Reranker reward.}
In our implementation, the reranker is a generative yes/no judge. Given an instruction $I$, source input $x$, and generated output $\hat{y}$, we define the semantic reward as the normalized probability of the affirmative answer:
\begin{equation}
\begin{aligned}
    z^+_\phi &= z_\phi(\texttt{yes} \mid I, x, \hat{y}), \\
    z^-_\phi &= z_\phi(\texttt{no} \mid I, x, \hat{y}), \\
    r_{\text{rank}}(x,\hat{y})
    &= \frac{\exp z^+_\phi}
    {\exp z^+_\phi + \exp z^-_\phi}.
\end{aligned}
\end{equation}
This score lies in $[0,1]$ and provides the main source-grounded semantic signal.

\paragraph{Reward safeguards.}
To reduce obvious degeneration, we combine the reranker score with the three safeguards shown in Figure~\ref{fig:sgsrl-overview}: a hard language gate, a batch-relative length penalty, and a repetition penalty. For a candidate $\hat{y}_i$ in reward batch $B$, the final reward is
\begin{equation}
\begin{aligned}
    s_i &= r_i^{\text{rank}}
    - \lambda_{\text{len}} p_i^{\text{len}}
    - \lambda_{\text{rep}} p_i^{\text{rep}}, \\
    r_i &= g_i \max(s_i,0).
\end{aligned}
\end{equation}
Here, $g_i$ is a hard target-language gate. In the Chinese--Thai setting, it requires the output to be predominantly Thai, contain no Chinese characters, and contain little Latin-script text. If the gate fails, the reward is set to zero.

The length penalty $p_i^{\text{len}}$ is computed with the reranker tokenizer and is relative to the current reward batch. It starts only when an output exceeds $2.5$ times the batch median length and increases linearly until $5.0$ times the median length. The repetition penalty $p_i^{\text{rep}}$ is based on repeated 4-grams and is activated when the repeated 4-gram ratio exceeds $0.15$. We set $\lambda_{\text{len}}=0.20$ and $\lambda_{\text{rep}}=0.05$ in the main configuration. Thus, semantic alignment is the only graded positive signal, while language validity, length control, and repetition control act as safeguards.

\paragraph{Policy optimization.}
We optimize the policy with GRPO~\citep{shao2024deepseekmath}. For each source input, we sample a group of candidate outputs, compute their rewards, normalize rewards within the group, and update the policy according to relative advantages. The reference policy is initialized from $\pi_{\theta_{\text{sft}}}$ to reduce drift from the cold-start target-language behavior.

We use this RL stage as semantic mid-training rather than final optimization. Directly optimizing the semantic reward can improve source-grounded coverage, but may also produce verbose or poorly formatted outputs. The purpose of this stage is therefore to learn source-to-target semantic alignment from $\mathcal{D}_{\text{src}}$, not to produce the final deployable generator.

\subsection{Target-form Recovery}

After semantic RL, we obtain an intermediate policy $\pi_{\theta_{\text{rl}}}$ that has absorbed source-side semantic supervision but may have degraded surface form. We then reuse $\mathcal{D}_{\text{para}}$ for 1 epoch of lightweight supervised recovery:
\begin{equation}
\begin{aligned}
    \theta_{\text{sg-srl}} =
    \arg\min_{\theta} \mathcal{L}_{\text{rec}}(\theta),
    \quad \theta \leftarrow \theta_{\text{rl}}, \\
    \mathcal{L}_{\text{rec}}(\theta)
    = - \mathbb{E}_{(x,y)\sim \mathcal{D}_{\text{para}}}
    \log \pi_{\theta}(y \mid x).
\end{aligned}
\end{equation}

Although initialization and recovery use the same parallel corpus, they serve different roles. Initialization teaches basic target-language generation from the base model, while recovery regularizes a semantically enhanced but form-degraded RL checkpoint. The final model keeps the semantic gains from source-grounded RL while restoring fluency, conciseness, and task format.

\section{Experiments}

We conduct a series of experiments to evaluate whether SG-SRL can use source-language monolingual data to improve low-resource target-language generation beyond cold-start SFT. The experiments are designed to answer four questions: (1) whether the full train--reinforce--recover pipeline improves target-language generation over SFT on a small parallel corpus, (2) why the intermediate semantic RL checkpoint should be treated as mid-training rather than the final model, (3) whether the learned source-grounded semantics transfer beyond the original title-generation format, and (4) whether the framework can generalize to a realistic low-resource language setting by replacing the LLM-based reranker with an encoder-based embedding reward when no strong target-language reranker is available.

\subsection{Experiment 1: Effectiveness of SG-SRL}

We first evaluate whether the full train--reinforce--recover pipeline improves target-language generation beyond cold-start SFT, testing the central claim that source-language monolingual data can provide useful semantic supervision when it is converted into a reference-free cross-lingual reward.

\paragraph{Task and data.}
The main task is Chinese-to-Thai news-title generation. We use CNewSum, a large-scale Chinese summarization dataset with human-annotated adequacy and deducibility levels \citep{10.1007/978-3-030-88480-2_31}, to construct a low-resource Chinese-to-Thai setting. We translate 15k Chinese titles into Thai. Among them, 10k parallel examples are used for target-form initialization and target-form recovery, and 5k examples are held out as the development set. We additionally sample 100k Chinese-only examples from CNewSum as source-language monolingual data for source-grounded semantic RL. The three splits are each deduplicated and mutually disjoint.
This setting matches the target scenario of SG-SRL: a small amount of target-language supervision is available for learning output form, while substantially more source-language data can be used for semantic mid-training.

\paragraph{Base model and training stages.}
All main Chinese-to-Thai experiments use SmolLM3-3B~\citep{bakouch2025smollm3} as the base model. The SG-SRL pipeline has three stages as described in Section~\ref{sec:method}. First, we train a cold-start supervised model on the 10k Chinese--Thai parallel examples. Second, we perform source-grounded semantic RL on the 100k Chinese-only examples. Third, we apply target-form recovery by fine-tuning the RL checkpoint for one epoch on the same 10k parallel examples. We refer to the supervised model after the first stage as \textsc{Cold-Start SFT}, and to the final recovered model as \textsc{SG-SRL}.


\paragraph{Evaluation protocol.}
For the main Chinese-to-Thai generation task, we use an LLM-judge protocol with \texttt{deepseek-v4-flash}. This choice reflects the metric-sensitivity issue of semantic-reward RL: source grounding, hallucination avoidance, and meaning-preserving rephrasings can be under-measured by surface-overlap or embedding-similarity metrics \citep{su2026semanticrl}; Appendix~\ref{app:metric-sensitivity} provides further discussion and explains why the reranker reward is not used as the main metric. The judge compares the gold Thai reference, \textsc{SG-SRL}, and \textsc{Cold-Start SFT} in terms of semantic adequacy, factual faithfulness, Thai fluency, and title-style conciseness. We report both three-way ranking statistics and direct pairwise win rates. For intermediate RL checkpoints, we additionally evaluate entity alignment, event alignment, factual consistency, fluency, and conciseness/format control, together with output length statistics to quantify verbosity.


\paragraph{Results.}
Table~\ref{tab:main-ranking} reports the three-way LLM-judge ranking among the gold Thai reference, \textsc{SG-SRL}, and \textsc{Cold-Start SFT} on the 5k CNewSum development examples. The total score assigns +2 points to Rank~1, +1 point to Rank~2, and -1 point to Rank~3. The gold reference remains the strongest candidate, showing that the task is still challenging. However, \textsc{SG-SRL} substantially improves over \textsc{Cold-Start SFT}: it is ranked first in 995 examples and second in 2602 examples, while \textsc{Cold-Start SFT} is ranked last in 3185 examples.

\begin{table*}[t]
\centering
\small
\setlength{\tabcolsep}{3pt}
\begin{subtable}[t]{0.59\textwidth}
\centering
\begin{tabular}{lccccc}
\hline
Model & Total Score & Avg. Score & Rank 1 & Rank 2 & Rank 3 \\
\hline
Gold & 7899 & 1.58 & 3723 & 865 & 412 \\
\textbf{\textsc{SG-SRL}} & \textbf{3189} & \textbf{0.64} & \textbf{995} & \textbf{2602} & \textbf{1403} \\
\textsc{Cold-Start SFT} & -1088 & -0.22 & 282 & 1533 & 3185 \\
\hline
\end{tabular}
\subcaption{Three-way ranking.}
\label{tab:main-ranking}
\end{subtable}
\hfill
\begin{subtable}[t]{0.38\textwidth}
\centering
\begin{tabular}{lcc}
\hline
Outcome & Count & Ratio \\
\hline
\textbf{\textsc{SG-SRL} wins} & \textbf{3613} & \textbf{72.3\%} \\
\textsc{Cold-Start SFT} wins & 1378 & 27.6\% \\
Tie & 9 & 0.2\% \\
\hline
\end{tabular}

\subcaption{Pairwise win rate.}
\label{tab:pairwise}
\end{subtable}
\caption{\textbf{Main effectiveness comparison on Chinese-to-Thai title generation.}\\
Left: three-way LLM-judge ranking among the gold reference, \textsc{SG-SRL}, and \textsc{Cold-Start SFT}; right: direct pairwise comparison between the two models. The exact judge prompts used for the three-way ranking and pairwise comparison are provided in Appendix~\ref{app:judge-prompts}.}
\label{tab:main-results}
\vspace{-0.5em}
\end{table*}

The direct pairwise comparison in Table~\ref{tab:pairwise} gives a more intuitive model-to-model comparison: \textsc{SG-SRL} wins against \textsc{Cold-Start SFT} in 72.3\% of examples, with very few ties. Together, these results indicate that SFT on a small parallel corpus can teach the model to produce Thai-form outputs, but it does not provide sufficient coverage for robust source-grounded generation. By contrast, SG-SRL uses the larger Chinese-only corpus to learn additional semantic grounding and then restores target-language form through recovery.

\paragraph{Analysis.}
The improvement is strongest in preference-based evaluation rather than in a narrow reference-matching setting, which is consistent with the goal of SG-SRL. The three-way comparison shows that the recovered RL model can sometimes compete even against the gold reference, while the pairwise comparison directly confirms its advantage over the supervised baseline. The method does not merely imitate a small set of Thai references; it uses Chinese-only inputs to strengthen source-grounded semantic coverage. The result supports the first part of our claim: semantic rewards can turn source-language monolingual data into effective supervision for low-resource target-language generation.

\subsection{Experiment 2: Semantic Learning versus Reward Hacking}
\label{sec:rl-diagnostics}

We next analyze the intermediate RL checkpoints before target-form recovery. This experiment asks why the semantic RL stage should be viewed as mid-training rather than as the final deployable generator. All variants start from the same \textsc{Cold-Start SFT} checkpoint and optimize on the same Chinese-only source corpus, but differ in reward design or RL control.

We compare four variants at a high level here, leaving the exact reward definitions and ablation rationale to Appendix~\ref{app:rl-reward-details}. \textsc{Gate+BatchLen} is the main SG-SRL reward configuration, combining a reranker semantic score with Thai-language gating, batch-relative length control, and repetition control. \textsc{Gate-Only} removes the explicit length and repetition safeguards. \textsc{RefLen} replaces batch-relative length control with source-conditioned absolute length control. Observing that recent analyses RL training suggest that improvements and failures can arise from the RL algorithm itself~\citep{liu2025understanding}, we introduce another setting \textsc{GRPO-Control}, where GRPO is replaced with Dr.GRPO~\citep{liu2025understanding} to examine whether verbosity and form degradation are mainly artifacts of the GRPO-style optimization, or the relevance-oriented reward.


\begin{table*}[t]
\centering
\small
\begin{tabular}{lccccccc}
\hline
Checkpoint & Entity & Event & Factual & Fluency & Conciseness & Avg. & Mean Length \\
\hline
\textbf{\textsc{Gate+BatchLen}} & \textbf{2.331} & \textbf{2.887} & \textbf{2.363} & \textbf{2.194} & \textbf{1.738} & \textbf{2.303} & \textbf{531.7} \\
\textsc{Gate-Only} & 1.777 & 2.166 & 1.776 & 1.583 & 1.219 & 1.704 & 866.6 \\
\textsc{RefLen} & 1.401 & 1.638 & 1.393 & 1.194 & 1.595 & 1.444 & 179.7 \\
\textsc{GRPO-Control} & 1.441 & 1.549 & 1.428 & 1.000 & 0.976 & 1.279 & 708.9 \\
\hline
\end{tabular}
\caption{\textbf{Semantic--form trade-off in intermediate RL checkpoints.} \textsc{Gate+BatchLen} achieves the strongest semantic alignment before recovery, but its outputs remain much longer than the gold Thai references, whose mean length is 162.1 characters. \textsc{Gate-Only} suffers from severe verbosity, while \textsc{RefLen} controls length but weakens semantic learning. A qualitative case study is provided in Appendix~\ref{app:rl-diagnostics-case}, and the five-dimensional judge prompt is provided in Table~\ref{tab:judge-prompt-five-dim}.}
\label{tab:rl-diagnostics}
\vspace{-1.2em}
\end{table*}

\paragraph{Results.}
Table~\ref{tab:rl-diagnostics} shows a clear semantic--form trade-off. \textsc{Gate+BatchLen} achieves the best overall score and the strongest entity, event, and factual alignment, indicating that source-grounded semantic RL can inject useful cross-lingual grounding into the model. However, its average output length is 531.7 characters, far longer than the gold Thai references, whose average length is 162.1 characters. Thus, even the best intermediate RL checkpoint is not directly deployable.

\paragraph{Analysis.}
The comparison with \textsc{Gate-Only} shows why auxiliary regularization is necessary. When the reward contains only the reranker score and the language gate, the model exploits the relevance reward by generating much longer outputs (average length 866.6 characters). \textsc{RefLen} shows the opposite failure mode. It reduces average output length to 179.7 characters, close to the gold reference length, but its entity, event, factual, and fluency scores drop substantially, suggesting that aggressive absolute length control can suppress verbosity but may also restrict the model's ability to explore and express source-side semantics in the target language.



The standalone \textsc{GRPO-Control} comparison further clarifies the cause of the failure. It obtains the lowest overall score while still producing long outputs, so the observed degeneration is not resolved by changing the RL control variant. The main issue is therefore not simply the GRPO-style optimizer, but the relevance-oriented reward structure itself.

These findings motivate the train--reinforce--recover design. The RL checkpoint is valuable because it learns source-grounded semantic alignment, but it should not be used as the final generator. Target-form recovery is needed to convert the semantically enhanced but form-degraded checkpoint into a usable target-language model.

\subsection{Experiment 3: Transfer Beyond Title Generation}
\label{sec:transfer}

We further test whether the semantic gains from SG-SRL transfer beyond the original title-generation format. This experiment asks whether SG-SRL learns reusable source-grounded semantics or only improves the specific training format.

\paragraph{Task and data.}
The transfer task is long-form Chinese-to-Thai translation, which requires preserving more entities, events, and factual relations over a longer context than title generation. We construct a 100-example long-form translation set with Chinese inputs longer than 500 characters. The data is constructed with GPT-5.5 assistance and manually checked. We also evaluate short-title translation, but report it in Appendix~\ref{app:short-translation} because the three models obtain very similar BLEU scores, making that setting less discriminative for semantic grounding.

\begin{table*}[t]
\centering
\small
\begin{tabular}{lccccccc}
\hline
Model & BLEU & Entity & Event & Factual & Thai Fluency & Completeness & Avg. \\
\hline
Base & 0.11 & 2.600 & 2.800 & 2.720 & 2.420 & 2.680 & 2.644 \\
\textsc{Cold-Start SFT} & 0.14 & 2.700 & 2.700 & 2.680 & 3.240 & 2.180 & 2.700 \\
\textbf{\textsc{SG-SRL}} & \textbf{0.17} & \textbf{3.420} & \textbf{3.540} & \textbf{3.520} & \textbf{3.280} & \textbf{3.260} & \textbf{3.404} \\
\hline
\end{tabular}
\caption{\textbf{Transfer to long-form Chinese-to-Thai translation.} BLEU \citep{papineni2002bleu} provides a surface-level automatic metric, while the LLM-judge dimensions evaluate semantic preservation and Thai quality. \textsc{SG-SRL} improves over both the base model and \textsc{Cold-Start SFT}, with the largest gains on entity alignment, event alignment, factual consistency, and completeness. The long-text translation judge prompt is provided in Table~\ref{tab:judge-prompt-long-translation}.}
\label{tab:long-form-transfer}
\vspace{-1.2em}
\end{table*}

\paragraph{Results.}
Table~\ref{tab:long-form-transfer} shows that \textsc{SG-SRL} improves over both the base model and \textsc{Cold-Start SFT} in BLEU. More importantly, the LLM-judge results show a clearer advantage across semantic dimensions. \textsc{SG-SRL} achieves the highest entity alignment, event alignment, factual consistency, and completeness while maintaining Thai fluency.

\paragraph{Analysis.}
The transfer results clarify the type of improvement learned by SG-SRL. The benefit is not a uniform gain on short translation instances; rather, it becomes more visible when the input requires longer-context semantic preservation and robust cross-lingual grounding. This supports the interpretation that source-grounded semantic RL improves reusable semantic alignment instead of only memorizing the title-generation format.

\subsection{Experiment 4: Language Generalization to Tibetan}
\label{sec:tibetan-embedding-reward}

Finally, we test whether SG-SRL can be instantiated in a target language that better matches a truly low-resource setting. Tibetan is a useful test case because it lacks a reliable LLM backbone that can be directly used as a cross-lingual reranker, unlike the Thai experiments where a stronger multilingual reranker is available. However, Tibetan is covered by CINO, a Chinese minority-language encoder model \citep{yang-etal-2022-cino}. We therefore replace the LLM-based reranker with a CINO-based Tibetan--Chinese embedding model and use its similarity score as the semantic reward.

\paragraph{Task and data.}
The task is Chinese-to-Tibetan generation with an embedding-based semantic reward. The data comes from the VLM portion of FTibSuite, a resource suite for Tibetan vision--language modeling \citep{anonymous2026ftibsuite}. This setting is fully in-domain: all data are drawn from a 100k-example Chinese--Tibetan parallel caption corpus, and the same corpus is used to train the embedding scorer and to construct the RL task. We split the corpus into 10k examples for cold-start supervised fine-tuning, 10k examples for development evaluation, and 80k examples for RL training.

The embedding model is trained on the Tibetan--Chinese caption pairs with a contrastive objective, mapping matched pairs closer in representation space and pushing mismatched pairs apart. Unlike a cross-encoder reranker, this model returns a similarity score in a shared embedding space. We compare two RL reward settings: one uses Tibetan text as the semantic reference, and the other uses Chinese text as the semantic reference. The comparison tests whether the learned embedding space can provide a usable cross-lingual reward when the semantic anchor is placed on either side of the bilingual pair.

\begin{table}[t]
\centering
\small
\begin{tabular}{lcc}
\hline
Setting & BLEU & Embedding Similarity \\
\hline
RL w/ Bo reference & 0.4519 & 0.7164 \\
RL w/ Ch reference & 0.4523 & 0.7011 \\
\hline
\end{tabular}
\caption{\textbf{Language generalization with embedding-based Tibetan rewards.} Using either Tibetan or Chinese as the semantic reference leads to comparable BLEU scores, suggesting that the trained encoder-based embedding model can provide an in-domain cross-lingual relevance signal when an LLM-based reranker is not available.}
\label{tab:tibetan-reward}
\vspace{-1.5em}
\end{table}

\paragraph{Results.}
As shown in Table~\ref{tab:tibetan-reward}, the two reference choices lead to similar BLEU scores. This result shows that the SG-SRL paradigm can still work when the reward module is implemented by an encoder-based embedding model rather than an LLM-based reranker. The Chinese-reference reward is especially relevant to SG-SRL, because it shows that target-language generation can be guided by a semantic signal anchored in the source language even when the target language lacks a strong reranker backbone.

\paragraph{Analysis.}
The Tibetan experiment should be interpreted together with the embedding-reward comparison summarized in Appendix~\ref{app:qwen-embedding-reward}. That appendix replaces the reranker reward in the Thai setting with \texttt{Qwen3-8B-Embedding} similarity and shows a clear trade-off: the embedding reward produces much shorter outputs and is less prone to length hacking, but its semantic supervision is weaker. In particular, candidate scores concentrate in a narrow 55--70 range, whereas the reranker separates good and bad generations more clearly and yields stronger post-recovery performance. Thus, generic embedding rewards are not a drop-in replacement for rerankers.

The Tibetan setting studies the complementary case where a strong reranker is unavailable, which is often the realistic constraint for genuinely low-resource languages. The CINO-based reward is effective because it is trained on the same Tibetan--Chinese caption domain used for RL, giving it sufficient in-domain resolution to distinguish matched from mismatched pairs. The comparable results with Tibetan and Chinese references further suggest that this learned cross-lingual space can support either target-anchored or source-anchored rewards. Therefore, Table~\ref{tab:tibetan-reward} shows that SG-SRL can be adapted with an in-domain encoder-based reward when no strong reranker exists, rather than claiming that embedding rewards generally replace rerankers or generalize broadly out of domain.


\section{Conclusion}
We presented SG-SRL, a novel \textbf{train-reinforce-recover framework} that enables \textbf{reference-free reinforcement learning} via cross-lingual semantic rewards to overcome the data bottleneck in low-resource target-language generation. Crucially, our decoupled approach isolates source-grounded semantic learning from target-form regularization, effectively neutralizing the \textbf{verbosity-based reward hacking} inherent in relevance optimization. Our evaluations across Chinese-to-Thai and Tibetan tasks, spanning both \textbf{cross-lingual summarization} and long-form translation, confirm that SG-SRL significantly enhances \textbf{semantic grounding and factual consistency} over standard SFT. This work establishes a robust pipeline for aligning weak target languages using high-resource source anchors, paving the way for more equitable multilingual model expansion.

\section*{Limitations}
A main limitation of this work is that SG-SRL is not yet fully matched to the most ideal low-resource setting. In principle, the framework benefits from a strong cross-lingual reranker that can directly score the semantic match between a source-language input and a target-language generation. However, such rerankers are usually unavailable for genuinely low-resource target languages. Therefore, while our Chinese-to-Thai experiments use a reranker-based reward, the Tibetan experiment uses an encoder-based embedding reward as a practical substitute. This shows that the framework can be adapted to a more realistic low-resource scenario, but it also means that the low-resource experiment does not exactly replicate the full reranker-based SG-SRL setting. Future work can further improve this part by building stronger reward models for low-resource languages or by designing reward functions that rely less on high-quality multilingual rerankers.

\bibliography{reference}

\appendix
\section{Additional Details}

\subsection{Metric Sensitivity}
\label{app:metric-sensitivity}

Semantic-reward RL should not be evaluated only with the same reward model used for training. In our setting, a reranker reward is useful as a training signal because it can compare a Chinese source input with a Thai generation without a gold Thai reference. However, using that same reranker as the main evaluation metric would risk overestimating models that exploit the reward by producing overly long outputs. We therefore report LLM-judge preference results for the main Chinese-to-Thai task and use BLEU only as a supplementary surface-overlap metric in transfer experiments.

\subsection{Reward Details for RL Diagnostics}
\label{app:rl-reward-details}

The intermediate-checkpoint comparison in Section~\ref{sec:rl-diagnostics} uses four reward/control variants. \textsc{Gate+BatchLen} is the main SG-SRL configuration: it combines the reranker relevance score with a Thai-language gate, a batch-relative length penalty, and a repeated 4-gram penalty. \textsc{Gate-Only} keeps the reranker score and Thai-language gate but removes explicit length and repetition penalties. \textsc{RefLen} replaces batch-relative length control with a source-conditioned absolute length constraint. \textsc{GRPO-Control} keeps the main reward form but changes the RL control variant to test whether degeneration is mainly caused by the optimization recipe rather than the relevance-oriented reward.

\subsection{Judge Prompts}
\label{app:judge-prompts}

For the main title-generation evaluation, the judge receives the Chinese source, the gold Thai title, and model outputs in randomized order. It ranks outputs by semantic adequacy, factual faithfulness, Thai fluency, and title-style conciseness. For pairwise evaluation, the judge compares \textsc{SG-SRL} and \textsc{Cold-Start SFT} directly and may return a tie only when the two outputs are indistinguishable in quality.

\begin{table*}[t]
\centering
\small
\begin{tabular}{p{0.18\textwidth}p{0.74\textwidth}}
\hline
Field & Prompt instruction \\
\hline
Input & Given a Chinese news input and a Thai candidate title, evaluate whether the Thai title preserves the source-side entities, events, and factual relations. \\
Entity & Score whether named entities, quantities, locations, organizations, and other salient participants are correctly preserved. \\
Event & Score whether the main action, event, or state described in the source is correctly expressed. \\
Factual & Score whether the output avoids unsupported claims, contradictions, and hallucinated details. \\
Fluency & Score whether the Thai output is grammatical, natural, and readable. \\
Conciseness & Score whether the output follows title style and avoids unnecessary verbosity or formatting artifacts. \\
Output & Return integer scores for the five dimensions and a brief justification. \\
\hline
\end{tabular}
\caption{Five-dimensional judge prompt used for intermediate RL-checkpoint diagnostics.}
\label{tab:judge-prompt-five-dim}
\end{table*}

\begin{table*}[t]
\centering
\small
\begin{tabular}{p{0.18\textwidth}p{0.74\textwidth}}
\hline
Field & Prompt instruction \\
\hline
Input & Given a long Chinese passage and a Thai translation, evaluate translation quality without relying only on word overlap. \\
Entity & Score whether important entities and numerical information are preserved. \\
Event & Score whether the major events, actions, and relations are translated correctly. \\
Factual & Score whether the translation is faithful to the source and avoids hallucination. \\
Thai fluency & Score whether the Thai text is fluent, grammatical, and coherent. \\
Completeness & Score whether the translation covers the important information in the source passage. \\
Output & Return dimension scores and an overall assessment. \\
\hline
\end{tabular}
\caption{Judge prompt used for long-form Chinese-to-Thai translation transfer evaluation.}
\label{tab:judge-prompt-long-translation}
\end{table*}

\subsection{Qualitative RL Diagnostic Case}
\label{app:rl-diagnostics-case}

The qualitative cases follow the same pattern as the aggregate results in Table~\ref{tab:rl-diagnostics}. \textsc{Gate-Only} outputs often mention more source-side content but become excessively long and sometimes drift away from title format. \textsc{RefLen} better controls length but frequently drops entities or event details. \textsc{Gate+BatchLen} provides the best compromise among the intermediate RL checkpoints, although it still requires target-form recovery before deployment.

\subsection{Short-Title Translation}
\label{app:short-translation}

We also evaluated a short-title translation setting. The task is less discriminative than long-form transfer because the source inputs contain fewer entities and event relations, and the compared models obtain similar surface-overlap scores. We therefore report the more informative long-form transfer results in the main text, where semantic preservation over longer contexts better reveals the effect of source-grounded semantic RL.

\subsection{Embedding Reward Comparison}
\label{app:qwen-embedding-reward}

As an alternative to the reranker reward, we tested a generic \texttt{Qwen3-8B-Embedding} similarity reward in the Thai setting. This reward is less prone to severe length hacking because longer outputs do not necessarily increase cosine similarity to a fixed semantic reference. However, it provides weaker score discrimination: candidate scores concentrate in a narrow range, whereas the reranker more clearly separates good and bad generations. This motivates the Tibetan experiment in Section~\ref{sec:tibetan-embedding-reward}, where an in-domain encoder-based reward is trained for the low-resource setting rather than directly substituting a generic embedding model for the reranker.

\end{document}